\pdfoutput=1
\documentclass{article}

\PassOptionsToPackage{round}{natbib}
\usepackage[preprint]{nips_2018_modify}

\usepackage[utf8]{inputenc} 
\usepackage[T1]{fontenc}    
\usepackage{hyperref}       
\usepackage{url}            
\usepackage{booktabs}       
\usepackage{amsfonts}       
\usepackage{nicefrac}       
\usepackage{microtype}      

\renewcommand{\cite}{\citep}
\usepackage{amsmath}
\usepackage{comment}
\usepackage{amssymb}
\usepackage{algorithm}
\usepackage{algorithmic}
\usepackage{graphicx} 
\usepackage{subcaption}
\usepackage{tabularx}

\renewcommand{\vec}[1]{\boldsymbol{#1}}
\newcommand{\vp}[2]{$\text{#1}\!\times\!\text{10}^{\text{#2}}$}
\newcommand{\vn}[2]{$\text{#1}\!\times\!\text{10}^{-\text{#2}}$}

\usepackage{color}

\title{Learning Graph Representation\\ via Formal Concept Analysis}

\author{
  Yuka Yoneda \\
  ISIR, Osaka University \\
  \texttt{yoneda@ar.sanken.} \\
  \texttt{osaka-u.ac.jp} \\
  \And
  Mahito Sugiyama \\
  National Institute of Informatics \\
  JST, PRESTO \\
  \texttt{mahito@nii.ac.jp} \\
  \And
  Takashi Washio \\
  ISIR, Osaka University \\
  \texttt{washio@ar.sanken.} \\
  \texttt{osaka-u.ac.jp} \\
}

\begin{document}

\maketitle

\begin{abstract}
We present a novel method that can learn a \emph{graph representation} from multivariate data.
In our representation, each node represents a cluster of data points and each edge represents the subset-superset relationship between clusters, which can be mutually overlapped.
The key to our method is to use \emph{formal concept analysis} (FCA), which can extract hierarchical relationships between clusters based on the algebraic closedness property.
We empirically show that our method can effectively extract hierarchical structures of clusters compared to the baseline method.
\end{abstract}

\section{Introduction}
\emph{Representation learning} has become one of the most important tasks in machine learning~\cite{Bengio13,Goodfellow16}.
The typical task is to find a numerical (vectorized) representation from structured objects, such as images~\cite{Krizhevsky12}, speeches~\cite{Graves13}, and texts~\cite{Mikolov13}, which have discrete structures between variables.
However, to date, learning of a structured representation from numerical data has not been studied at sufficient depth,
while the task is crucial for \emph{relational reasoning} aiming at finding relationships between objects to bridge between a symbolic approach and a gradient-based numerical approach~\cite{Santoro17}.

A classic yet promising branch of research for learning structures from numerical data is \emph{hierarchical clustering}~\cite{hie_clu}, which is a widely used unsupervised learning method in multivariate data analysis from natural language processing~\cite{PeterF} to human motion analysis~\cite{FengZ}.
Given a set of data points without any class labels, hierarchical clustering can learn a tree structured representation, called a \emph{dendrogram}, whose nodes correspond to clusters and leaves correspond to the input data points.
The resulting tree structures can be used for further analysis of relational reasoning and other machine learning tasks.

However, the representation in hierarchical clustering is restricted to the form of a \emph{binary tree} since clusters must be disjoint with each other in the exiting approaches.
Nevertheless, clusters of objects can often overlap in real-world data analysis.
Therefore the technique that can find a hierarchical structure of overlapped clusters from multivariate data is needed, which leads to a more flexible \emph{graph structured representation} of numerical data.

To solve the problem, we propose to combine {\it nearest neighbor-based binarization} and {\it formal concept analysis} (FCA)~\cite{FCA}. 
FCA can extract a hierarchical structure of data using the algebraic closedness property, which consists of mutually overlapped clusters~\cite{Valtchev04}.
Since FCA is designed for binary data, we first binarize numerical data by nearest neighbor-based binarization, which can model local geometric relationships between data points. 

The remainder of this paper is organized as follows. 
Section~\ref{propose_section} introduces our method; Section~\ref{knn_section} explains nearest neighbor based binarization and Section~\ref{fca_section} introduces FCA.
Section~\ref{experi_section} empirically examines our method and Section~\ref{conc_section} summarizes our contribution.

\section{The Proposed Method}\label{propose_section}
We introduce our method that learns a graph representation from numerical data.
It consists of two stages.
It first binarizes numerical data by nearest neighbor-based binarization, followed by applying formal concept analysis (FCA)~\cite{FCA} to the binarized data, which is an established method to analyze relational databases~\cite{Kaytoue10}.
Input to our method is an unlabeled real-valued vectors.
Let $D = \{\vec{d}_1, \vec{d}_1, \dots, \vec{d}_n\}$ be an input dataset.
Each data point is an $m$-dimensional vector and is denoted by $\boldsymbol{d}_i = (d_i^1, d_i^2, \dots, d_i^m)\in \mathbb{R}^m$.

\subsection{Nearest Neighbor-based Binarization}\label{knn_section}
In the first stage, we convert each $m$-dimensional vector $\vec{d}_i \in \mathbb{R}^m$ into an $n$-dimensional binary vector $\vec{z}_i \in \{0,1\}^n$, where $n$ coincides with the number of data points.
The $j$th feature in the converted binary vector $\vec{z}_i$ shows whether or not the $j$th data point $\vec{d}_j$ belongs to nearest neighbors of $\vec{d}_i$.
Formally, given a dataset $D \subset \mathbb{R}^m$ and a parameter $k \in \mathbb{N}$, each data point $\vec{d}_i \in D$ is binarized to the binary vector $\boldsymbol{z}_i = (z_i^1, z_i^2, \dots , z_i^n) \in \{0, 1\}^n$, where each component $z_i^j$ is defined as
\begin{equation}\label{z_defi}
\begin{split}
z_i^j=\left\{
\begin{array}{ll}
1 &\text{if } \boldsymbol{d}_j \text{ is the }l\text{th nearest data point from } \boldsymbol{d}_i \text{ for } l \leq k,\\
0 &\text{otherwise}.
\end{array}
\right.
\end{split}
\end{equation}
Hence our binarization models local relationships between data points in terms of the relative closeness in the original feature space, which is often used as $k$-nearest neighbor graphs in spectral clustering~\cite{vonluxburg07}.

When we regard indices of data points as \emph{items}, every binary vector $\boldsymbol{z}_i \in \{0, 1\}^n$ can be directly treated as a \emph{transaction} $X_i \subseteq \{1, 2, \dots, n\}$ defined as $X_i = \{\,j \in \{1,2,\dots n\} \mid z_i^j = 1\,\}$.
In other words, $X_i$ is the set of indices of the data points which are the $l$th nearest data points ($l \leq k$) from $\boldsymbol{d}_i$, and hence $|X_i| = k$ always holds.
Output in this stage is the transaction database $\mathcal{T} = \{X_1, X_2, \dots, X_n\}$.
Transaction databases are the standard data format in frequent pattern mining~\cite{freq_pattern} and other fields in databases.

\subsection{Formal Concept Analysis}\label{fca_section}
In the second stage, we apply formal concept analysis (FCA)~\cite{FCA} to the transaction database obtained from the first stage,
which is a mathematical way to analyze databases based on the lattice theory and can be viewed as a co-clustering method for binary data.
FCA can obtain hierarchical relationships of the original numerical data via the binarized transaction database.

Let $\mathcal{A} \subseteq \mathcal{T}$ be a subset of transactions and $B \subseteq [n] = \{1,2,\dots,n\}$ be a subset of data indices. 
We define that $\mathcal{A}'$ is the set of indices common to all transactions in $\mathcal{A}$ and $B'$ is the set of transactions possessing all indices in $B$; that is,
\begin{align*}
 \mathcal{A}' = \{\,j \in [n] \mid j \in X_i \text{ for all }X_i \in \mathcal{A} \,\}, \quad
 B' = \{\,X_i \in \mathcal{T} \mid B \subseteq X_i \,\}. 
\end{align*}
The pair $(\mathcal{A}, B)$ is called a {\it concept} if and only if $\mathcal{A}' = B$ and $B' = \mathcal{A}$. 
Here the mapping $''$ is a \emph{closure operator}, as it satisfies $\mathcal{A} \subseteq \mathcal{A}''$, $\mathcal{A} \subseteq \mathcal{C} \Rightarrow \mathcal{A}'' \subseteq \mathcal{C}''$, and $(\mathcal{A}'')'' = \mathcal{A}''$,
and $\mathcal{A}$ is \emph{closed} if and only if $(\mathcal{A}, B)$ is a concept.
A concept $(\mathcal{A}_1, B_1)$ is less general than a concept $(\mathcal{A}_2, B_2)$ if $\mathcal{A}_1$ is contained in $\mathcal{A}_2$; that is, 
 $(\mathcal{A}_1, B_1) \leq (\mathcal{A}_1, B_1) \Longleftrightarrow \mathcal{A}_1 \subseteq \mathcal{A}_2$,
where the relation ``$\le$'' becomes a partial order.
The {\it concept lattice} is the set of concepts equipped with the order $\le$. 
Intuitively, concepts are representative clusters in the dataset.

\begin{figure}[t]
 \centering
 \includegraphics[width=.9\linewidth]{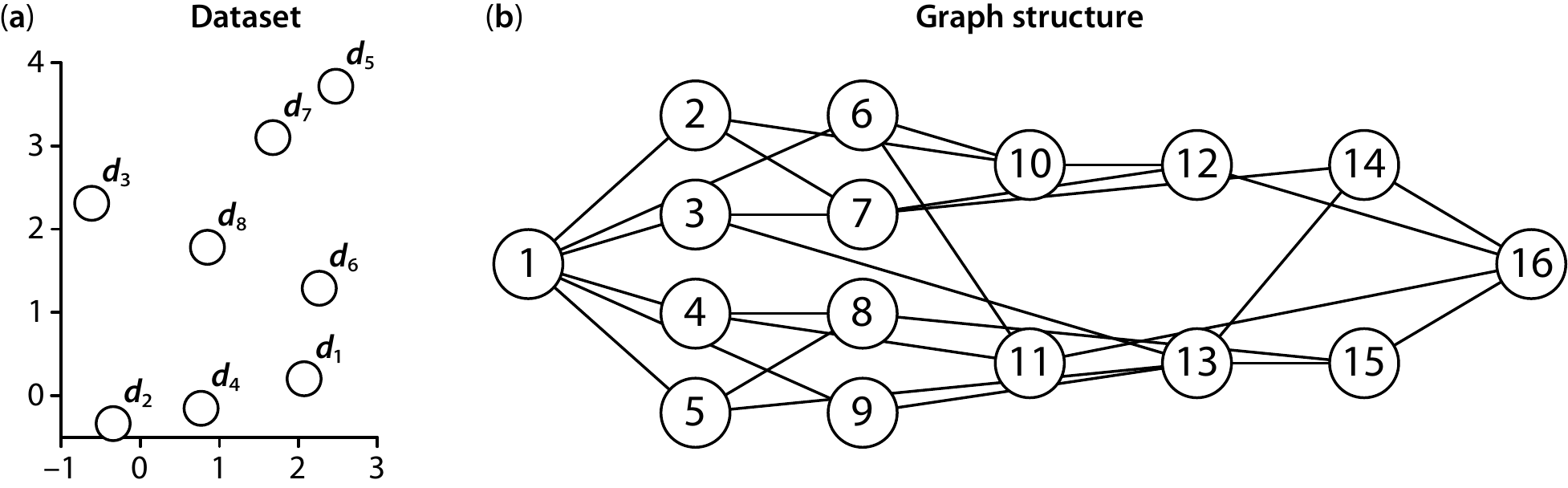}
 \caption{(\textbf{a}) Example of dataset. (\textbf{b}) Graph representation obtained from the dataset with $k = 3$. All edges are directed from left to right.
 Numbers of nodes indicate clusters as follows:
 1: $\emptyset$, 2: $\{\vec{d}_1\}$, 3: $\{\vec{d}_6\}$, 4: $\{\vec{d}_3\}$, 5: $\{\vec{d}_8\}$, 6: $\{\vec{d}_2,\vec{d}_4\}$, 7: $\{\vec{d}_1,\vec{d}_6\}$, 8: $\{\vec{d}_3,\vec{d}_8\}$, 9: $\{\vec{d}_5,\vec{d}_7\}$, 10: $\{\vec{d}_1,\vec{d}_2,\vec{d}_4\}$, 11: $\{\vec{d}_2,\vec{d}_3,\vec{d}_4\}$, 12: $\{\vec{d}_1,\vec{d}_2,\vec{d}_4,\vec{d}_6\}$, 13: $\{\vec{d}_5,\vec{d}_6,\vec{d}_7,\vec{d}_8\}$, 14: $\{\vec{d}_1,\vec{d}_5,\vec{d}_6,\vec{d}_7,\vec{d}_8\}$, 15: $\{\vec{d}_3,\vec{d}_5,\vec{d}_6,\vec{d}_7,\vec{d}_8\}$, 16: $\{\vec{d}_1,\vec{d}_2,\vec{d}_3,\vec{d}_4,\vec{d}_5,\vec{d}_6,\vec{d}_7,\vec{d}_8\}$.}
 \label{hasse}
\end{figure}

From the set $\mathfrak{L}$ of concepts, we finally construct a graph representation $G = (V, E)$, where $V = \{S \subseteq D \mid (T(S), T(S)') \in \mathfrak{L}\}$ with $T(S) = \{X_i \in \mathcal{T} \mid \vec{d}_i \in S\}$ and a directed edge $(v, w) \in E$ exists if $v$ covers $w$; that is, $v \subset w$ and $v \subseteq u \subset w \Rightarrow u = v$.
Hence $v \subseteq w$ if and only if $w$ is reachable from $v$.
This graph coincides with the Hasse diagram of the concept lattice using the partial order $\le$.
We illustrate an example of a graph representation in Figure~\ref{hasse}(\textbf{b}) obtained by our method from a dataset in Figure~\ref{hasse}(\textbf{a}).

Since the set of concepts is equivalent to that of \emph{closed itemsets} used in closed itemset mining~\cite{lattice_closed},
we can efficiently enumerate all concepts using a closed itemset mining algorithm such as LCM~\cite{lcm}.
Moreover, we can directly obtain more compact representations by pruning nodes with small clusters using \emph{frequent closet itemset mining} as the frequency (or the support) of an itemset coincides with the size of a cluster.

\section{Experiments}\label{experi_section}
We evaluate the proposed method using synthetic and real-world datasets.
Since our method can be viewed as hierarchical clustering, to assess the effectiveness of our method,
we compare our method with the standard hierarchical agglomerative clustering (HAC) with Ward's method~\cite{ward_original}.

We performed all experiments on Windows10 Pro 64bit OS with a single processor of Intel Core i7-4790 CPU 3.60 GHz and 16GB of main memory. 
All experiments were conducted in Python 3.5.2.
In our method, we used LCM~\cite{lcm} version 5.3\footnote{\url{http://research.nii.ac.jp/~uno/code/lcm53.zip}} for closed itemset mining.
HAC is implemented in scipy~\cite{scipy}.

We use dendrogram purity (DP)~\cite{dp_first} for evaluation.
Dendrogram purity is the standard measure to evaluate the quality of hierarchical clusters~\cite{dp_use}.
Given a dataset $D = \{\boldsymbol{d}_1, \boldsymbol{d}_2, \dots, \boldsymbol{d}_n\}$ and its ground truth partition $\mathcal{C} = \{C_1, C_2, \dots, C_P\}$ such that $\bigcup_{C_i \in \{1, \dots, P\}} C_i = D$ and $C_i \cap~C_j = \emptyset$, and let $\mathcal{H}$ be a set of clusters obtained by a hierarchical clustering algorithm.
We denote by $\mathrm{LCA}(\boldsymbol{d}_i, \boldsymbol{d}_j) \in \mathcal{H}$ the smallest cluster that includes both $\boldsymbol{d}_i$, $\boldsymbol{d}_j$ in $\mathcal{H}$ and $\mathrm{pur}(F, G) = |F \cap G| / |F|$ for a pair of clusters $F, G \subseteq D$.
Assume that $Q$ be the pair of data points in the same cluster; that is,
$Q = \{(\boldsymbol{d}_i, \boldsymbol{d}_j) \mid \boldsymbol{d}_i, \boldsymbol{d}_j \in C_l \text{ for some } C_l \in \mathcal{C}\}$.
The \emph{dendrogram purity} of hierarchical clusters $\mathcal{H}$ is defined as
\begin{align*}
 DP(\mathcal{H}) = \frac{1}{|Q|}\sum\nolimits_{l=1}^P \sum\nolimits_{\boldsymbol{d}_i, \boldsymbol{d}_j \in C_{l}} \mathrm{pur}(\mathrm{LCA}(\boldsymbol{d}_i, \boldsymbol{d}_j), C_{l}).
\end{align*}
The dendrogram purity takes values from $0$ to $1$ and larger is better.

We use three types of synthetic datasets \texttt{synth1}, \texttt{synth2}, and \texttt{synth3}.
\texttt{synth1} consists of two equal sized clusters sampled from two normal distributions
$(\mu_0, \sigma^2_0) = (0, 1)$ and $(\mu_1, \sigma^2_1) = (2, 1)$ for each feature.
\texttt{synth2} consists of two equal sized clusters sampled from two normal distributions
$(\mu_0, \sigma^2_0) = (0, 1)$ and $(\mu_1, \sigma^2_1) = (2, 4)$ for each feature.
\texttt{synth3} consists of three clusters with the size ratio $(2, 1, 1)$ sampled from three two-dimensional multivariate normal distributions, where
the mean is randomly sampled from $[-25, 25]$ and the variance is always $1$ for each feature.
For each dataset, we obtained the averaged dendrogram purity from 10 trials.

\setlength{\tabcolsep}{5pt}
\begin{table}[t]
 \begin{small}
 \centering
 \caption{Experimental results, where $c$ denotes the number of classes.}
 \begin{tabularx}{1.0\linewidth}{lrrXrrrrll} \toprule
  Name & $n$ & $m$ & $c$ & \multicolumn{2}{c}{\# clusters} & \multicolumn{2}{c}{DP} & \multicolumn{2}{c}{Runtime (sec.)}\\
  &&&& Ours & HAC & Ours & HAC & Ours & HAC\\ \midrule
  \texttt{synth1} & 100 & 2 & 2 & 77,364.5 & 199 & \textbf{0.937} & 0.812 & \vp{1.24}{0} & \vn{1.01}{3}\\
  \texttt{synth1\_large} & 1,000 & 500 & 2 & 58.4 & 1,999 & \textbf{1.0} & \textbf{1.0} & \vp{1.42}{1} & \vn{4.52}{1}\\
  \texttt{synth2} & 100 & 2 & 2 & 27,445.3 & 199 & \textbf{0.842} & 0.705 & \vn{5.88}{1} & \vn{9.02}{4}\\
  \texttt{synth3} & 100 & 2 & 3 & 425.2 & 199 & \textbf{0.976} & 0.936 & \vn{1.64}{1} & \vn{9.12}{4}\\ \midrule
  \texttt{parkinsons} & 197 & 23 & 2 & 263,189 & 393 & \textbf{0.828} & 0.738 & \vp{1.30}{1} & \vn{5.32}{3}\\
  \texttt{vertebral} & 310 & 6 & 2 & 503,476,064 & 619 & \textbf{0.872} & 0.686 & \vp{2.55}{4} & \vn{4.25}{3}\\
  \texttt{breast\_cancer} & 569 & 10 & 2 & 3,142 & 1,137 & \textbf{0.869} & 0.771 & \vp{3.40}{0} & \vn{8.43}{3}\\
  \texttt{wine\_red} & 1,600 & 12 & 2 & 24,412,834 & 3,199 & \textbf{0.849} & 0.845 & \vp{3.73}{3} & \vn{8.09}{2}\\
  \texttt{ctg} & 2,126 & 20 & 2 & 1,426,981 & 4,251 & \textbf{0.800} & 0.765 & \vp{2.52}{2} & \vn{1.23}{1}\\
  \texttt{seismic\_bumps} & 2,584 & 25 & 2 & 91,059 & 5,167 & 0.931 & \textbf{0.943} & \vp{9.48}{1} & \vn{1.52}{1}\\ \bottomrule
 \end{tabularx}
 \label{realdatasets}
 \end{small}
\end{table}

\begin{figure}[t]
\centering
 \includegraphics[width=.9\linewidth]{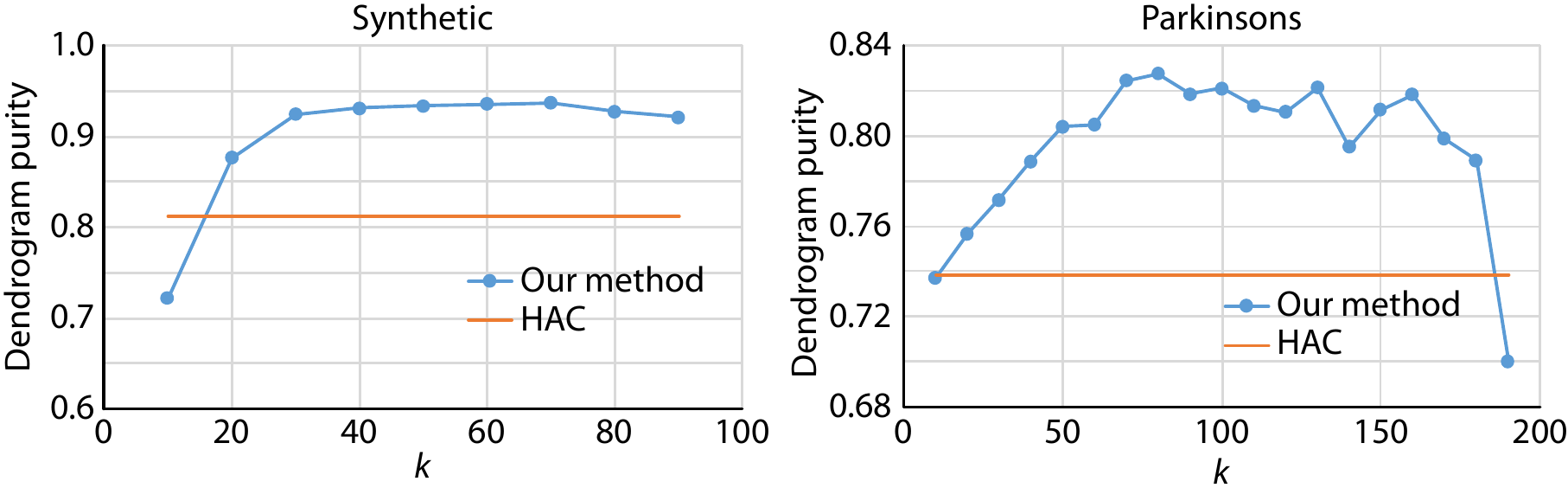}
 \caption{Result on \texttt{synth1} (left) and \texttt{parkinsons} (right).}
 \vskip -10pt
 \label{result_da1_mean}
\end{figure}

We collected six real-world datasets from UCI machine learning repository~\cite{UCI} and used only continuous features.
The statistics of datasets are summarized in Table~\ref{realdatasets}. 

\subsection{Results and Discussion}\label{result_dis_subsection}
First we examine the sensitivity of our method with respect to the parameter $k$ for $k$-nearest neighbor binarization using the synthetic dataset \texttt{synth1} with $n = 100$ and $m = 2$ and the real-world dataset \texttt{parkinsons} with $n = 197$ and $m = 23$.
We plot results in Figure \ref{result_da1_mean}, where we varied $k$ from 10 to 90 for \texttt{synth1} and from 10 to 190 for \texttt{parkinsons}.
It shows that when $k \ge 20$, the dendrogram purity is higher than HAC in both datasets and it is stable for larger $k$ except for $k = 190$ in \texttt{parkinsons}, which is almost the same as the dataset size.
This means that our method is robust to changes in $k$ if $k$ is set to be sufficiently large.
In the following, we set $k$ to be the half of the respective dataset size.

Next we examine the clustering performance of our method compared to HAC across various types of datasets.
Results are summarized in Table~\ref{realdatasets}.
To prune unnecessary small clusters in our method, we set the lower bound of the size of clusters as $190$ for \texttt{ctg}, \texttt{seismic\_bumps}, and $490$ for \texttt{synth1\_large}.
They clearly show that our method is consistently superior to HAC across all synthetic and real-world datasets except for \texttt{seismic\_bumps}.
The reason is that our method can learn overlapped clusters while HAC cannot.
Although the number of clusters in HAC is always fixed to $2n - 1$ as it learns a binary tree, our method allows more flexible clustering, resulting in a larger number of clusters as shown in Table~\ref{realdatasets}.
How to effectively use the lower bound of the size of clusters to reduce clusters is our future work.

To summarize, our results show that the proposed method is robust to the parameter setting and can obtain better quality hierarchical structures than the standard baseline, hierarchical agglomerative clustering with Wald's method.
This means that a graph representation learned by our method can be effective for further data analysis for relational reasoning.

\section{Conclusions}\label{conc_section}
In this paper, we have proposed a novel method that can learn graph structured representation of numerical data.
Our method first binarizes a given dataset based on nearest neighbor search and then applies formal concept analysis (FCA) to the binarized data.
The extracted concept lattice corresponds to a hierarchy of clusters, which leads to a directed graph representation.
We have experimentally showed that our method can obtain more accurate hierarchical clusters compared to the standard hierarchical agglomerative clustering with Wald's method.

\textbf{Acknowledgments}:
This work was supported by JSPS KAKENHI Grant Numbers JP16K16115, JP16H02870, and JST, PRESTO Grant Number JPMJPR1855, Japan (M.S.);
and JSPS KAKENHI Grant Number 15H05711 (T.W.).


\begin{thebibliography}{21}
\providecommand{\natexlab}[1]{#1}
\providecommand{\url}[1]{\texttt{#1}}
\expandafter\ifx\csname urlstyle\endcsname\relax
  \providecommand{\doi}[1]{doi: #1}\else
  \providecommand{\doi}{doi: \begingroup \urlstyle{rm}\Url}\fi

\bibitem[Aggarwal and Han(2014)]{freq_pattern}
C.~C. Aggarwal and J.~Han, editors.
\newblock \emph{Frequent Pattern Mining}.
\newblock Springer, 2014.

\bibitem[Bengio et~al.(2013)Bengio, Courville, and Vincent]{Bengio13}
Y.~Bengio, A.~Courville, and P.~Vincent.
\newblock Representation learning: {A} review and new perspectives.
\newblock \emph{IEEE Transactions on Pattern Analysis and Machine
  Intelligence}, 35\penalty0 (8):\penalty0 1798--1828, 2013.

\bibitem[Brown et~al.(1992)Brown, deSouza, Mercer, Pietra, and Lai]{PeterF}
P.~F. Brown, P.~V. deSouza, R.~L. Mercer, V.~J.~D. Pietra, and J.~C. Lai.
\newblock Class-based \emph{n}-gram models of natural language.
\newblock \emph{Computational Linguistics}, 18\penalty0 (4):\penalty0 467--479,
  1992.

\bibitem[Davey and Priestley(2002)]{FCA}
B.~A. Davey and H.~A. Priestley.
\newblock \emph{Introduction to Lattices and Order}.
\newblock Cambridge University Press, 2 edition, 2002.

\bibitem[Goodfellow et~al.(2016)Goodfellow, Bengio, and
  Courville]{Goodfellow16}
I.~Goodfellow, Y.~Bengio, and A.~Courville.
\newblock \emph{Deep Learning}.
\newblock MIT Press, 2016.

\bibitem[Graves et~al.(2013)Graves, Mohamed, and Hinton]{Graves13}
A.~Graves, A.~Mohamed, and G.~Hinton.
\newblock Speech recognition with deep recurrent neural networks.
\newblock In \emph{2013 IEEE International Conference on Acoustics, Speech and
  Signal Processing}, pages 6645--6649. IEEE, 2013.

\bibitem[Heller and Ghahramani(2005)]{dp_first}
K.~A. Heller and Z.~Ghahramani.
\newblock Bayesian hierarchical clustering.
\newblock In \emph{Proceedings of the 22nd International Conference on Machine
  Learning}, pages 297--304, 2005.

\bibitem[Jones et~al.(2001--)Jones, Oliphant, Peterson, et~al.]{scipy}
Eric Jones, Travis Oliphant, Pearu Peterson, et~al.
\newblock {SciPy}: Open source scientific tools for {Python}, 2001--.
\newblock URL \url{http://www.scipy.org/}.
\newblock [Online; accessed <today>].

\bibitem[Kaytoue et~al.(2011)Kaytoue, Kuznetsov, Napoli, and
  Duplessis]{Kaytoue10}
M.~Kaytoue, S.~O. Kuznetsov, A.~Napoli, and S.~Duplessis.
\newblock Mining gene expression data with pattern structures in formal concept
  analysis.
\newblock \emph{Information Sciences}, 181:\penalty0 1989--2001, 2011.

\bibitem[Kobren et~al.(2017)Kobren, Monath, Krishnamurthy, and
  McCallum]{dp_use}
A.~Kobren, N.~Monath, A.~Krishnamurthy, and A.~McCallum.
\newblock A hierarchical algorithm for extreme clustering.
\newblock In \emph{Proceedings of the 23rd ACM SIGKDD International Conference
  on Knowledge Discovery and Data Mining}, pages 255--264, 2017.

\bibitem[Krizhevsky et~al.(2012)Krizhevsky, Sutskever, and
  Hinton]{Krizhevsky12}
A.~Krizhevsky, I.~Sutskever, and G.~E. Hinton.
\newblock {ImageNet} classification with deep convolutional neural networks.
\newblock In \emph{Advances in Neural Information Processing Systems 25}, pages
  1097--1105, 2012.

\bibitem[Lichman(2013)]{UCI}
M.~Lichman.
\newblock {UCI} machine learning repository, 2013.

\bibitem[Maimon and Rokach(2005)]{hie_clu}
O.~Maimon and L.~Rokach, editors.
\newblock \emph{Data Mining and Knowledge Discovery Handbook}.
\newblock Springer, 2005.

\bibitem[Mikolov et~al.(2013)Mikolov, Sutskever, Chen, Corrado, and
  Dean]{Mikolov13}
T.~Mikolov, I.~Sutskever, K.~Chen, G.~S. Corrado, and J.~Dean.
\newblock Distributed representations of words and phrases and their
  compositionality.
\newblock In \emph{Advances in Neural Information Processing Systems 26}, pages
  3111--3119, 2013.

\bibitem[Pasquier et~al.(1999)Pasquier, Bastide, Taouil, and
  Lakhal]{lattice_closed}
N.~Pasquier, Y.~Bastide, R.~Taouil, and L.~Lakhal.
\newblock Efficient mining of association rules using closed itemset lattices.
\newblock \emph{Information Systems}, 24\penalty0 (1):\penalty0 25--46, 1999.

\bibitem[Santoro et~al.(2017)Santoro, Raposo, Barrett, Malinowski, Pascanu,
  Battaglia, and Lillicrap]{Santoro17}
A.~Santoro, D.~Raposo, D.~G.~T. Barrett, M.~Malinowski, R.~Pascanu,
  P.~Battaglia, and T.~Lillicrap.
\newblock A simple neural network module for relational reasoning.
\newblock \emph{arXiv:1706.01427}, 2017.

\bibitem[Uno et~al.(2004)Uno, Asai, Uchida, and Arimura]{lcm}
T.~Uno, T.~Asai, Y.~Uchida, and H.~Arimura.
\newblock An efficient algorithm for enumerating closed patterns in transaction
  databases.
\newblock In \emph{Discovery Science}, volume 3245 of \emph{LNCS}, pages
  16--31, 2004.

\bibitem[Valtchev et~al.(2004)Valtchev, Missaoui, and Godin]{Valtchev04}
P.~Valtchev, R.~Missaoui, and R.~Godin.
\newblock Formal concept analysis for knowledge discovery and data mining: The
  new challenges.
\newblock In \emph{Concept Lattices}, volume 2961 of \emph{LNCS}, pages
  352--371, 2004.

\bibitem[von Luxburg(2007)]{vonluxburg07}
U.~von Luxburg.
\newblock A tutorial on spectral clustering.
\newblock \emph{Statistics and Computing}, 17\penalty0 (4):\penalty0 395--416,
  2007.

\bibitem[Ward~Jr(1963)]{ward_original}
J.~H. Ward~Jr.
\newblock Hierarchical grouping to optimize an objective function.
\newblock \emph{Journal of the American Statistical Association}, 58\penalty0
  (301):\penalty0 236--244, 1963.

\bibitem[Zhou et~al.(2013)Zhou, Torre, and Hodgins]{FengZ}
F.~Zhou, F.~D.~l. Torre, and J.~K. Hodgins.
\newblock Hierarchical aligned cluster analysis for temporal clustering of
  human motion.
\newblock \emph{IEEE Transactions on Pattern Analysis and Machine
  Intelligence}, 35\penalty0 (3):\penalty0 582--596, 2013.

\end{thebibliography}

\end{document}